%% file: main.tex
\documentclass[10pt,twocolumn,letterpaper]{article}

\usepackage{iccv}
\usepackage{times}
\usepackage{epsfig}
\usepackage{graphicx}
\usepackage{amsmath}
\usepackage{amssymb}

\usepackage{pifont}
\usepackage{xcolor}
\usepackage{multirow}
\usepackage{siunitx}

\definecolor{magenta}{rgb}{1,0.5,0}

\newcommand*\colourcheck[1]{%
  \expandafter\newcommand\csname #1check\endcsname{\textcolor{#1}{\ding{51}}}%
}
\colourcheck{green}

\newcommand*\colourcross[1]{%
  \expandafter\newcommand\csname #1cross\endcsname{\textcolor{#1}{\ding{55}}}%
}
\colourcross{red}
\colourcross{magenta}
\colourcross{orange}

\usepackage[pagebackref=true,breaklinks=true,letterpaper=true,colorlinks,bookmarks=false]{hyperref}

\iccvfinalcopy 

\newif\ifarxiv

\def\iccvPaperID{****} 

\ificcvfinal\fi
\begin{document}
\arxivtrue
\title{6-DOF GraspNet: Variational Grasp Generation for Object Manipulation}

\author{Arsalan Mousavian\\
NVIDIA\\
{\tt\small amousavian@nvidia.com}
\and
Clemens Eppner\\
NVIDIA\\
{\tt\small ceppner@nvidia.com}
\and
Dieter Fox\\
NVIDIA\\
{\tt\small dieterf@nvidia.com}
}

\maketitle

\newcommand{\arsalan}[1]{\textcolor{red}{{\bf Arsalan: #1}}}
\newcommand{\clemens}[1]{\textcolor{green}{{\bf Clemens: #1}}}
\newcommand{\dieter}[1]{\textcolor{magenta}{{\bf Dieter: #1}}}

\begin{abstract}
Generating grasp poses is a crucial component for any robot object manipulation task. In this work, we formulate the problem of grasp generation as sampling a set of grasps using a variational autoencoder and assess and refine the sampled grasps using a grasp evaluator model.  Both Grasp Sampler and Grasp Refinement networks take 3D point clouds observed by a depth camera as input. We evaluate our approach in simulation and real-world robot experiments. Our approach achieves 88\% success rate on various commonly used objects with diverse appearances, scales, and weights. Our model is trained purely in simulation and works in the real world without any extra steps. The video of our experiments can be found {\href{https://research.nvidia.com/publication/2019-10_6-DOF-GraspNet\%3A-Variational}{here}}.
\end{abstract}

\input{01_intro.tex}
\input{02_related_works.tex}
\input{03_method.tex}
\input{04_experiments.tex}
\input{05_conclusions.tex}

{\small
\bibliographystyle{ieee_fullname}
\bibliography{main}
}

\end{document}

%% file: 01_intro.tex
\section{Introduction}

\begin{figure}
    \centering
    \includegraphics[width=0.48\textwidth]{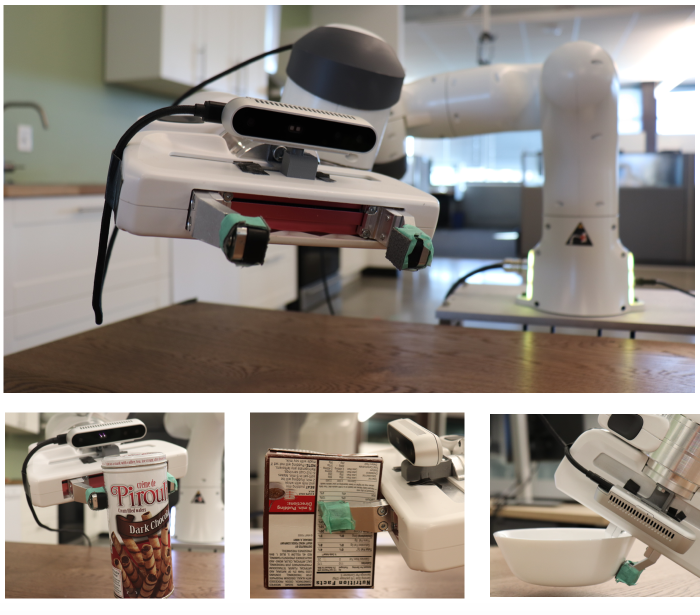}
    \caption{The Franka Panda manipulator used in our experiments. Our approach is able to efficiently generate diverse sets of grasps that lead to successful pickups of unknown objects. }
    \label{fig:title_image}
\end{figure}

\begin{figure*}
    \centering
    \begin{tabular}{ccc}
         \includegraphics[height=0.20\textwidth]{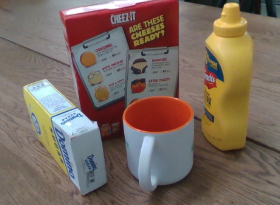} &
         \includegraphics[height=0.20\textwidth]{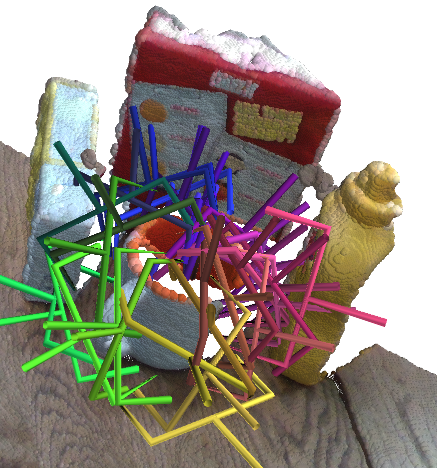} &
         \includegraphics[height=0.20\textwidth]{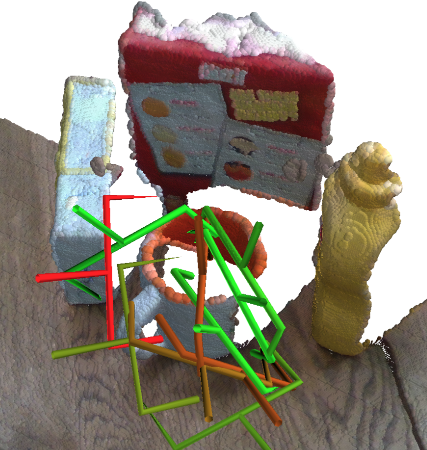}
    \end{tabular}
    \caption{Visualization of the predicted grasps for the mug. (middle) All the grasps that are generated by our method. (right) Grasps that are both kinematically feasible and collision free color-coded by the predicted scores. Green is the highest and red is the lowest.}
    \label{fig:grasp_pruning}
    
\end{figure*}

Grasp selection is one of the most important problems in robot manipulation. Here, a robot observes an object and needs to decide where to move its gripper (3D position and 3D orientation) in order to pickup the object (see Fig.~\ref{fig:title_image}). Grasp selection is complex since the stability of grasps depends on object and gripper geometry, object mass distribution, and surface frictions. The geometry around an object poses additional constraints on which grasp points are reachable without causing the robot manipulator to collide with other objects in a scene (see Fig.~\ref{fig:grasp_pruning}). Typically, this problem is approached by geometry-inspired heuristics to select promising grasp points around an object, possibly followed by a more in-depth geometric analysis of the stability and reachability of a sampled grasp~\cite{ten2017grasp}. Many of these approaches rely on the availability of complete 3D models of an object, which is a severe limitation in realistic scenarios where a robot only observes a scene with a noisy depth camera, for example. 
To overcome this limitation, one could move the camera to generate a full object model or perform shape completion, followed by geometry-based grasp analysis. However, moving the camera might be impossible in constrained spaces, and shape completion might not be sufficiently accurate for grasp generation and evaluation.

Recently, several groups have introduced deep learning techniques to evaluate the quality of grasps from raw point cloud data~\cite{pinto2016supersizing,mahler2017dex,ten2017grasp,liang2018pointnetgpd}.  While these approaches provide good grasp assessments, they still use manually designed heuristics to sample grasps for evaluation or rely on black-box optimization techniques such as CEM~\cite{mahler2017dex,yan2017learning}. Additionally, they do not provide efficient means for improving sampled grasps. In this paper, we introduce the first learning-based framework for efficiently generating diverse sets of stable grasps for unknown objects.  Our approach introduces two network architectures that sample, evaluate, and improve grasps. The key contributions of this paper are:
\begin{itemize}
    \item A variational auto-encoder (VAE) that can be trained to map the partial point cloud of an observed object to a diverse set of grasps for the object. Importantly, our VAE provides high coverage of all possible, functioning grasps while generating only a small number of failing grasps.  
    \item To improve the precision of the VAE samples, we introduce a grasp evaluator network that maps a point cloud of the observed object and the robot gripper to a quality assessment of the 6D gripper pose. Crucially, we show that the gradient of this network can be used to improve grasp samples, for instance moving gripper out of collision or ensuring that the gripper is well aligned with the object. 
    \item We demonstrate that our approach outperforms previous approaches and enables a robot to pickup 17 objects with a success rate of 88\%. Generating diverse grasps is quite important because not all the grasps are kinematically feasible for the robot to execute. We furthermore show that our approach generates diverse sets of grasp samples while maintaining high success rate. 
\end{itemize}

The paper is organized as follows. We first contrast related approaches to grasping that use deep learning, and then explain the different components of our approach: grasp sampling, evaluation, and refinement. Finally, we evaluate our method on a real robotic platform and show the effect of different hyperparameters in various ablation studies.

%% file: 02_related_works.tex
\section{Related Work}

\paragraph{Learning 6-DOF Grasps}
The prevailing approaches to solve the robot grasping problem are data-driven~\cite{bohg2014data}. While earlier methods were based on hand-crafted feature vectors~\cite{saxena2008learning,bohg2010learning,herzog2012template}, recent methods exploit convolutional architectures to operate on raw visual measurements~\cite{len20115deep,redmon2015real,pinto2016supersizing,mahler2017dex,levine2017handeye}.
Most of these grasp synthesis approaches are enabled by representing the grasp as an oriented rectangle in the image~\cite{jiang2011rectangle}. This 3-DOF representation constrains the gripper pose to be parallel to the image plane. The drawbacks of such a representation are manifold: Since it limits the grasp diversity, picking up an object might be impossible given additional constraints imposed by the arm or task. In case of a static image sensor it also leads to a severely restricted workspace~\cite{mahler2017dex}.

Our approach tackles the problem of predicting the full 6-DOF pregrasp pose. This is challenging due to occluded object parts that affect grasp success.
Yan et al.~\cite{yan2017learning} circumvent this problem by including the auxiliary task of reconstructing the geometry of the target object. The main task of predicting the 6-DOF grasp outcome can then use local geometry that is not part of the measurement.
Similar to our evaluator network, Zhou et al.~\cite{zhou20176dof} learn a grasp score function which they also use for grasp refinement. In contrast to our approach, both methods~\cite{yan2017learning,zhou20176dof} are only evaluated in simulation.
Similar to our grasp refinement phase, Lu et al.~\cite{lu2018planning} use the gradient of a learned grasp success model to infer the maximum likelihood grasp estimate.

Few methods formulate the problem as a regression to a single best grasp pose~\cite{schmidt2018grasping,min2019highdof}.
They inherently lack the ability to predict a diverse distribution of possible grasps.
Choi et al.~\cite{choi2018learning} classify 24 pre-defined orientations to chose a 6-DOF pre-grasp pose.
Such a coarse resolution of $SO(3)$ will necessarily lead to a limited diversity of the predicted grasps.
In contrast, the grasp point detection method~(GPD)~\cite{ten2017grasp,liang2018pointnetgpd} uses a more dense sampling of candidate grasps: A point in the observed point cloud is sampled randomly and a Darboux frame is constructed which is aligned with the estimated surface normal and the local direction of the principal curvature. Although this heuristic creates a quite diverse set of candidate grasps, it fails generating grasps along thin structures such as rims of mugs, plates, or bowls since estimating those surface normals from noisy measurements is challenging. Our learned grasp sampler does not suffer from such bias. As a result our proposed method finds grasps where GPD is not able to~(see Sec.~\ref{sec:real_experiments}).

Apart from using supervised learning, grasping has also been formulated as a reinforcement learning problem~\cite{pmlr-v87-kalashnikov18a,zeng2018learning} or approximations of it~\cite{levine2017handeye}.
The learned grasp policies are more expressive than describing only the final grasp pose.
Still, the action space of these methods is usually $se(2)$, limiting the diversity to top-down grasps.



\paragraph{Deep Neural Networks for Learning from 3D Data} The success of deep learning on 3D point cloud data started much later than its huge success on RGB images. In the early days, 3D data were represented as 3D voxels~\cite{Maturana-2015-6018} or as extracting features from 2.5 depth images~\cite{guptaECCV14} and process them similar to RGB image using convolutional neural networks which oftentimes lead to marginal improvements. Qi et al.~\cite{qi2016pointnet,qi2017pointnetplusplus} introduced a new architecture, called PointNet and PointNet++, that is capable of representing the 3D data and extract the representation efficiently. The success of PointNet lead to the introduction of different variations of network architectures~\cite{dgcnn, su18splatnet} that represent 3D data, showing significant improvement on 3D~object pose estimation, semantic segmentation, and part segmentation~\cite{su18splatnet,qi2017pointnetplusplus,qi2017frustum, danfei_pf}. In order to estimate a successful gasp, the 6-DOF pose of the grasps needs to be accurate. Operating on a single RGB image does not provide the required accuracy since the input and output are not in the same domain.
Therefore, we use 3D point clouds and PointNet++~\cite{qi2017pointnetplusplus} to generate and evaluate grasps in $SE(3)$.

\paragraph{Variational Autoencoders} Variational autoencoders~\cite{vae_first}~(VAE) are one of the main categories of deep generative models. VAEs can be trained in an unsupervised manner to maximize the likelihood of the training data. They have been applied to a variety of tasks such as future prediction~\cite{vae_desire, vae_prediction}, generating novel view points~\cite{NIPS2015_5851} and object segmentation~\cite{cvae}. In this work, we use a VAE to sample a diverse set of grasps in $SE(3)$.

The overall architecture of our model is similar to GANs~\cite{GAN}. The generator module is a VAE that is based on different samples from a latent space and the observed point cloud $X$.
It generates different grasp proposals and the evaluation network (discriminator) accepts or rejects them based on how likely it is that they are successful. Both generator and discriminator are taking the 3D point cloud $X$ of the object as part of the input.

%% file: 03_method.tex
    


\begin{figure*}[ht]
    \centering
    \begin{tabular}{cc}
         \includegraphics[height=3.7cm]{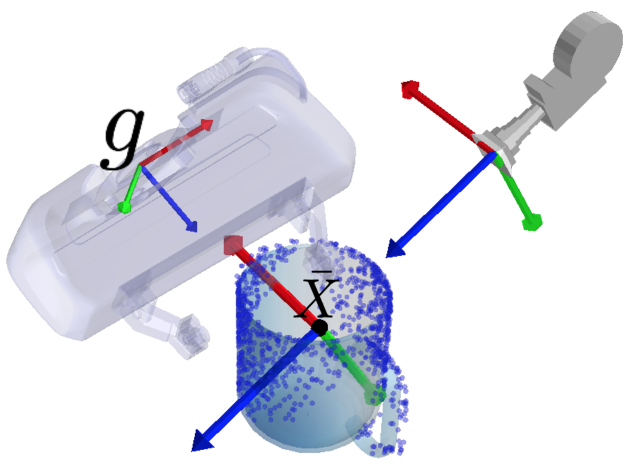}& 
         \includegraphics[height=3.7cm]{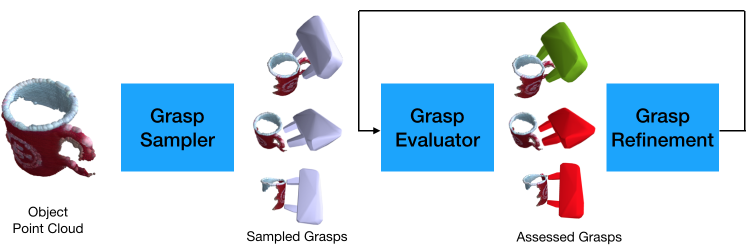}\\
         Grasp coordinate frame & Overview of our method
    \end{tabular}
    \caption{(left) Grasps are estimated with respect to the center of mass of the object point cloud, $X$. The axes of the grasp coordinate frame are parallel to those of the camera. (right) The object point cloud~$X$ is extracted from a depth image using plane fitting. The Grasp Sampler Network takes the point cloud and proposes different grasps. The evaluator network assesses the grasps based on the object point cloud and the proposed grasp. Grasps are improved iteratively using the gradient of the evaluator network }
    \label{fig:overview}
\end{figure*}

\section{6-DOF Grasp Pose Generation}

We formulate grasp pose generation as the process of producing sets of robot gripper poses such that closing the gripper at any of these poses results in a stable grasp of an object.  Furthermore, the process should generate \emph{diverse} sets of poses that ultimately cover all possible ways an object could be grasped. Robot gripper poses are given in $SE(3)$, specifying the 3D translation and 3D orientation of the gripper. Here, we focus on generating grasp poses for single objects, additional constraints due to a manipulator's reach and due to other objects in a scene are beyond the scope of this work and can be handled by trajectory optimization techniques. Grasp pose generation is challenging due to the narrow subspace of successful grasps in the space of all possible grasps. Small perturbations in the pose of a grasp can transform a successful grasp into a failure. To generate diverse sets of stable grasps, our approach samples grasp poses using a variational auto-encoder network followed by an iterative evaluation and refinement process. The input to our approach is a point cloud of the object the robot should pickup.  




Specifically, we aim to learn the posterior distribution~$P(G^* \mid X)$,  where $G^*$ represents the space of all successful grasps and $X$ is the partial point cloud of the object observed by a camera. Each grasp~$g \in G^*$ is represented by~$(R, T) \in SE(3)$ where $R \in SO(3)$ and $T \in \mathbb{R}^3$ are the rotation and translation of grasp~$g$. Grasps are defined in the object reference frame, whose origin is~$\bar{X}$, the center of mass of the observed point cloud. Its axes are parallel to those of the camera frame~(see Fig.~\ref{fig:overview}-a). The distribution of successful grasps~$G^*$ can be complex and discontinuous. For example, the distribution of $G^*$ for a mug has multiple modes along the rim, handle, and bottom. Within each mode, the space of successful grasps is continuous but grasps of different modes can be separated from each other. The total number of separate modes for each object category varies based on the shape and scale of objects.

Since the number of modes of~$G^*$ is not known beforehand, we propose to learn a generator module that maximizes the likelihood of successful grasps~$g \in G^*$. Since the generator only observes successful grasps during training, it is possible that it also generates failed grasps~$g \in G^-$. In order to detect and refine these negative grasps, an evaluation module is trained to predict~$P(S \mid g, X)$, i.e., the probability of success for a grasp~$g$ and the observed point cloud~$X$. Applied to a sampled grasp, the evaluation module predicts grasp success and propagates the success gradient back through the network to generate an improved grasp pose. This process can be repeated. Discarding all grasps that remain below a threshold provides the final set of high quality grasps. The overview of our method is shown in Fig.~\ref{fig:overview}-b.


\subsection{Variational Grasp Sampler}

\begin{figure}
    \includegraphics[width=0.5\textwidth]{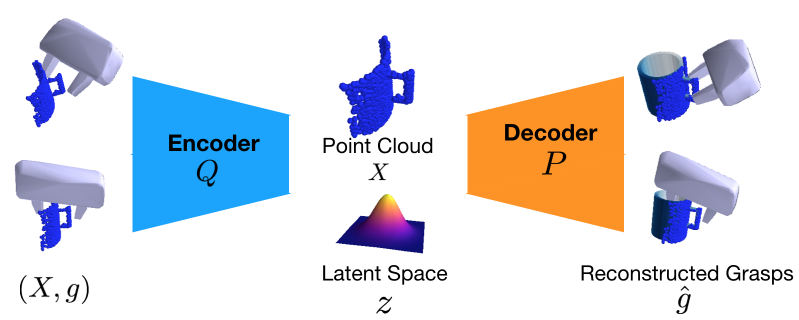}
    \caption{During training, the encoder maps each grasp to a point $z$ in a latent space. The distribution of the latent space is minimized toward a normal distribution. The decoder takes the point cloud and latent values and reconstructs the 6D grasps, visualized here as gripper poses. }
    \label{fig:vae}
\end{figure}

The grasp sampler, shown in Fig.~\ref{fig:vae}, is a generative model that maximizes~$P(G \mid X)$, the likelihood of a set of pre-defined successful grasps~$g \in G^*$. Given a point cloud~$X$ and a latent variable~$z$, the sampler is a deterministic function that predicts a grasp. It is assumed that~$P(z)$, the probability density function of the latent space, is known and chosen beforehand. In our approach, we use $P(z) = \mathcal{N}(0, I)$. Given a point cloud~$X$, different grasps are generated by sampling different~$z$ from $P(z)$. The likelihood of the generated grasps can be written as follows: 
\begin{equation}
\label{eq:vae}
P(G \mid X) = \int P(G \mid X, z; \Theta) P(z) dz
\end{equation}
Optimizing Eq.~\eqref{eq:vae} for each positive grasp $g \in G^*$ requires to integrate over all the values of the latent space, which is intractable. In order to make Eq.~\eqref{eq:vae} tractable, the encoder $Q(z \mid X, g)$ maps each pair of point cloud $X$ and grasp $g$ to a small subspace in the latent space $z$. Given the sampled $z \sim Q$, the decoder reconstructs the grasp $\hat{g}$. During training, the encoder and decoder are optimized to minimize the reconstruction loss $\mathcal{L}(g, \hat{g})$ between ground truth grasps $g \in G^*$ and the reconstructed grasps $\hat{g}$. Furthermore, the KL-divergence $\mathcal{D}_{KL}$ between the distribution $Q(\cdot|\cdot)$ and the normal distribution $\mathcal{N}(0, I)$ is minimized to ensure a normal distributed latent space with unit variance. The loss function is defined as follows: 
\begin{equation}
\label{eq:vae_loss}
\mathcal{L}_{\text{vae}} = \sum_{z \sim Q, g \sim G^*} \mathcal{\mathcal{L}}(\hat{g}, g)  - \alpha \mathcal{D}_{KL} \left[Q(z|X, g), \mathcal{N}(0, I)\right]
\end{equation}
Eq.~\eqref{eq:vae_loss} is optimized using stochastic gradient descent. For each mini-batch, the point cloud~$X$ is sampled for an object observed from a random viewpoint. For the sampled point cloud~$X$, grasps~$g$ are sampled from the set of ground truth grasps~$G^*$ using stratified sampling. To combine the orientation and translation loss, we define the reconstruction loss as follows:
\begin{equation}
\label{eq:recon_loss}
    \mathcal{L}(g, \hat{g}) = \frac{1}{n} \sum ||\mathcal{T}(g; p) - \mathcal{T}(\hat{g}; p)||_1
\end{equation}
where $\mathcal{T}(\cdot; p)$ is the transformation of a set of predefined points $p$ on the robot gripper. During training, the decoder learns to decode the latent value~$z$ that is sampled from $\mathcal{N}(0, I)$ and generates grasps while the encoder learns to output~$z$ such that it contains enough information to reconstruct the grasp pose while maintaining the normal distribution. During inference, the encoder~$Q$ is removed and latent values are sampled from~$N(0, I)$.

Both encoder and decoder are based on the PointNet++~\cite{qi2017pointnetplusplus} architecture. In this architecture, each point has a 3D coordinate and a feature vector. The features at each layer are computed based on the features of each point and the 3D relation of the points with respect to each other. The features of each input point~$x \in X$ are concatenated to~$g = [R, T]$. In the decoder, each point feature is concatenated with the latent variable $z$. The encoder learns to  compress the relative information of point cloud $X$ and latent variable grasp $g$ in such a way that it can be reconstructed by the decoder. 
\ifarxiv
Fig.~\ref{fig:latent_coloring} qualitatively shows that the latent space has strong correlation with grasp pose.

\begin{figure}
    \centering
    \begin{tabular}{cc}
         \includegraphics[width=0.22\textwidth]{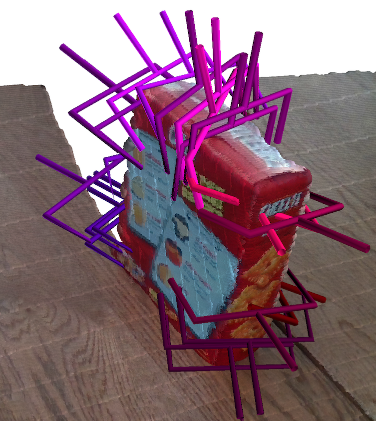}& \includegraphics[width=0.22\textwidth]{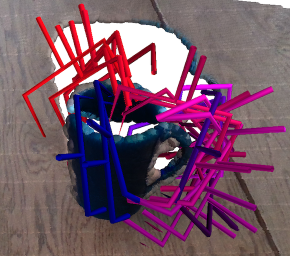}
    \end{tabular}
    \caption{Relation between grasp pose and corresponding latent value: Each grasp is colored by the corresponding latent value. Red and blue channels are set to the normalized latent value and green channel is set to 0. The smoothness of color transition between each grasp shows the strong correlation between the latent space and the grasp pose.}
    \label{fig:latent_coloring}
\end{figure}
\fi

\subsection{Grasp Pose Evaluation}
The grasp sampler trains the continuous posterior distribution~$P(G \mid X, z)$ using only positive grasps. As a result, it might contain failed grasps that are in between the modes of the distribution. These transitional grasps and other false positives need to be identified and pruned out. To do so, we need a grasp evaluation network that assigns a probability of success $P(S|g, X)$ to each grasp. This network needs to reason about grasps relative to the observed point cloud $X$, but it must also be able to extrapolate to unobserved parts of the object. Other methods learn to classify grasps based only on local observed parts of an object~\cite{ten2017grasp,mahler2017dex}.   In practice, the observed point cloud of an object has imperfections such as missing or noisy depth values. To mitigate this problem, previous methods resort to use high quality depth sensors \cite{mahler2017dex} or using multiple views \cite{ten2017grasp} which limits the deployment of the system outside of controlled environments. In this work, we classify each grasp using only the imperfect observed point cloud $X$ of the object.

Success of a grasp pose depends on the relative pose of the grasp with respect to the object. The inputs to the evaluator network are point cloud $X$ and grasp~$g$. Similar to the Grasp Sampler, we use the PointNet~\cite{qi2016pointnet} architecture for the Grasp Evaluator. There are multiple ways for classifying grasps. The first, simple approach is to associate the 6D pose of the grasp~$g$ to the features of each point $x \in X$ in the first layer. Our experiments showed that such a representation leads to poor accuracy in grasp classification. Instead, we propose to represent a grasp~$g$ in a way more closely tied to the object point cloud: We approximate the robot gripper by a point cloud, $X_g$, rendered according to the 6D grasp pose~$g$.   The object point cloud~$X$ and gripper point cloud~$X_g$ are combined into a single point cloud by using an extra binary feature that indicates whether a point belongs to the object or to the gripper. In the PointNet architecture, the features for each point are functions of features of the point itself and its neighbors plus the relative spatial relation of the points. Using the unified point cloud~$X \cup X_g$ makes it natural to use all the relative information between grasp pose~$g$ and object point cloud~$X$ for classifying the grasps. The grasp evaluator is optimized using the cross-entropy loss by optimizing
\begin{equation}
\label{eq:success_loss}
\mathcal{L}_{\text{evaluator}} = -\left( y log(s) + (1-y)log(1-s) \right)
\end{equation}
where $y$ is the ground truth binary label of the grasps indicating whether the grasp is successful or not and $s$ is the predicted probability of success by the evaluator.

In order to train a robust evaluator, the model needs to be trained with both positive and negative grasps.
Since the space of all possible 6D grasp poses is combinatorially large, it is not possible to sample all the negative grasps. Instead, we do hard negative mining to sample negative grasps. The set of hard negative grasps $G^-$ is defined as the grasps that have similar pose to a positive grasp but that are either in collision with the object or are too far from the object to grasp the object. More formally $G^-$ is defined as:
\begin{equation}
\label{eq:hard_negative}
    G^- = \{ g^- \mid \exists g \in G^* : \mathcal{L}(g, g^-) < \epsilon \}
\end{equation} 
where $\mathcal{L}(\cdot,\cdot)$ is defined in Eq.~\eqref{eq:recon_loss}.
During training, $g^-$ is sampled from a set of pre-generated negative grasps and by randomly perturbing positive grasps to make the mesh of the gripper either collide with the object mesh or to move the gripper mesh far from the object.

\begin{figure}
    \centering
    \begin{tabular}{cc}
         \includegraphics[width=0.22\textwidth]{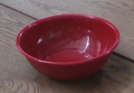}& \includegraphics[width=0.22\textwidth]{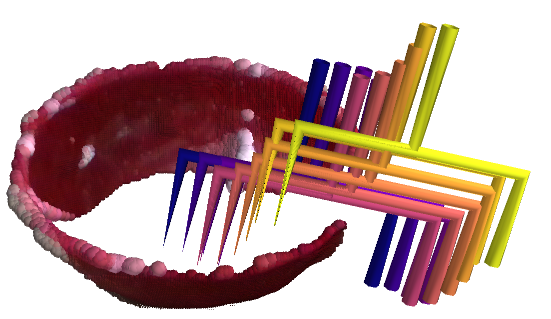} \\
         \includegraphics[height=0.22\textwidth]{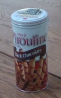}
         & \includegraphics[height=0.22\textwidth]{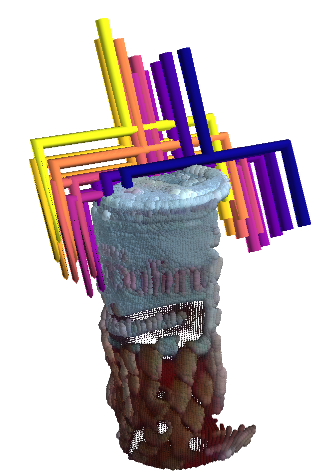}
    \end{tabular}
    \caption{Iterative Grasp Refinement: 
      (left) Image of the object. (right) Grasps colored according to refinement iteration. Dark blue are grasps initially generated from the VAE and yellow are final, refined grasps. Note that even though there are no points between the gripper fingers for the initial bowl grasp (blue), the evaluation network is able to push the gripper to a successful grasp pose. 
    }
    \label{fig:refinement}
\end{figure}
\subsection{Iterative Grasp Pose Refinement} 
Although the evaluation network rejects implausible grasps, a large portion of the rejected grasps can be close to successful ones. This insight can be exploited by searching for a transformation~$\Delta g \in SE(3)$ that turns an unsuccessful grasp into a successful one. More formally, we are looking for a refining transformation~$\Delta g$ that increases the probability of success, i.e., $P(s=1 \mid g + \Delta g) > P(s=1 \mid g)$. The evaluation network represents a differentiable function of success~$s$ based on the point cloud~$X$ and grasp~$g$. The refinement transformation that leads to maximum improvement in success probability can be computed by taking the derivative of success with respect to the grasp transformation: $\partial S / \partial g$. The partial derivative~$\partial S / \partial g$ provides the transformation for each point in the gripper point cloud~$X_g$ so as to increase the probability of success. Since the derivative is computed with respect to each point on the gripper independently, it can lead to non-rigid transformations for~$X_g$. To enforce the rigidity constraint, the transformed gripper point cloud~$X_g$ is defined as a function of orientation of the grasp defined in Euler angles $R_g = (\alpha_g, \beta_g, \gamma_g)$ and translation $T_g$. Using the chain rule, $\Delta g$ is computed as follows:
\begin{equation}
    \Delta g = \frac{\partial S}{\partial g} = \eta \times \frac{\partial S}{\partial \mathcal{T}(g; p)} \times  \frac{\partial \mathcal{T}(g; p)}{\partial g}
\end{equation}

Since the partial derivative $\partial S / \partial g$ is only a valid approximation in the local neighborhood, we use hyper-parameter $\eta$ to limit the magnitude of updates at each step. In practice, we choose $\eta$ in such a way that the maximum translation update of the grasp does not exceed~$\SI{1}{\cm}$. Fig.~\ref{fig:refinement} shows the refinement of estimated grasps at different iterations.

%% file: 04_experiments.tex
\section{Experiments}
\begin{figure}
    \centering
    \includegraphics[height=3cm]{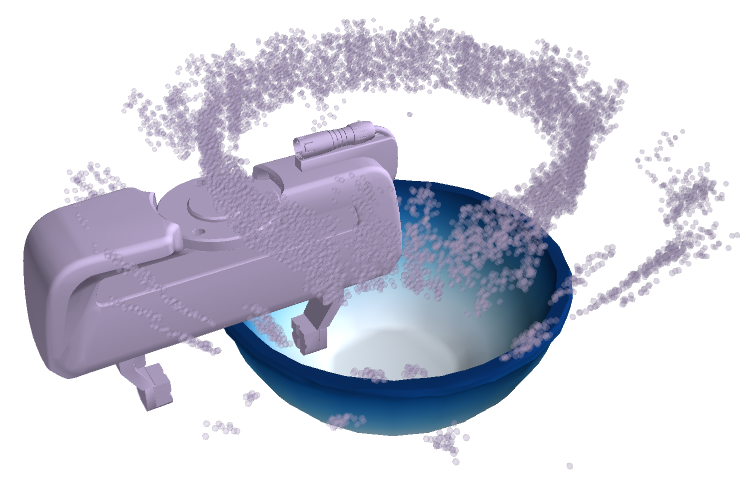}\hfill%
    \includegraphics[height=3cm]{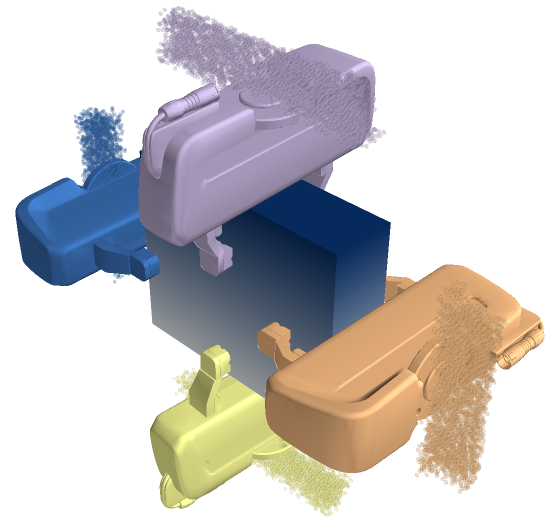}
    \caption{We use training data generated with a physics simulator. The colored dots around the objects depict successful grasps for a bowl~(left) and a box~(right). For each continuous grasp subspace an exemplary gripper pose is shown.}
    \label{fig:simulation_data}
\end{figure}

\paragraph{Training Data for Grasping}
\label{sec:flex}
To generate reference sets of successful grasps, we use the physics simulation FleX~\cite{macklin2014unified}, which provides realistic simulation of grasps for arbitrary object shapes.
Candidate grasps are sampled based on the object geometry. We sample random points on the object mesh surface and align the gripper's z-axis~(see Fig.~\ref{fig:overview}-a) with the surface normal. The distance between the gripper and the object surface is sampled uniformly between zero and the gripper's finger length. The orientation around the z-axis is also drawn from a uniform distribution. We only simulate grasps that are not in collision and whose closing volume between the fingers intersects the object. 
In total we use 206 objects from six categories in ShapeNet~\cite{chang2015shapenet}: boxes and cylinders (randomly generated), as well as bowls, bottles and mugs. A total of 10,816,720 candidate grasps are sampled of which we simulate 7,074,038 (65.4\%), i.e., those that pass the non-empty closing volume test. The simulation consists of a free-floating parallel-jaw gripper and the free-floating object without gravity~(similar to~\cite{zhou20176dof}). Surface friction and object density are kept constant. After closing its fingers the gripper executes a predefined shaking motion. A grasp is labeled successful if the object is kept between both fingers.
Overall, we generate 2,104,894 successful grasps (19.4\%). The resulting positive grasp labels are densely distributed as shown in the examples in Fig.~\ref{fig:simulation_data}.

\ifarxiv
\else
\vspace{-5mm}
\fi
\paragraph{Training}
Both grasp generator and evaluator networks are using PointNet++\cite{qi2017pointnetplusplus} and have similar architectures. Both modules consist of three set-abstraction layers followed by fully connected layers.
Each batch of the training data for the generator network consists of a rendering of the object from a random view and 64~grasps that are sampled using stratified sampling to make sure the sampled grasps have enough diversity. The weight for KL-Divergence loss~($\alpha$ in Eq.~\eqref{eq:vae_loss}) is set to~$0.01$. Each batch of training data for the evaluator network consists of~$30\%$ positive, $30\%$~negative grasps, and $40\%$~hard negative grasps. Hard negative grasps are selected from perturbed positive grasps by applying $\pm 0.6$ radians in each axis and $\pm 3cm$ to translation.
Both models are trained with the Adam optimizer using a learning rate of $0.0001$. All the grasps are generated in simulation and no real data was used to train any of the models (see Sec. \ref{sec:flex}). 


\ifarxiv
\paragraph{Network Architecture Details} Both grasp generator and evaluator are based on PointNet++ architecture. Both models consists of three set abstraction layers. Each set abstraction layer samples 128, 32, and all the points. Each set abstraction layer samples the points that are within \SI{2}{\cm}, \SI{4}{\cm}, and $\infty$ radius of the sampled points. Each set abstraction layer uses 3 fully connected layers to compute the features. The number of channels for each set abstraction layer are $[64,64,128]$, $[128,128,256]$ and $[256,256,512]$ respectively. The set abstraction layers are followed by two fully connected layers with $1024$ units. The grasp generator network outputs a rotation~$R$ that is represented as unit quaternion and a translation~$T$. Quaternions are generated by applying L2-Norm on a linear fully connected layer. No normalization is done for the translation~$T$. The evaluator network predicts the score for each grasp using a softmax layer. 
\fi

\ifarxiv
\else
\vspace{-5mm}
\fi
\paragraph{Evaluation Metrics}
\label{sec:metrics}
 We used two metrics to quantitatively evaluate grasping methods: \emph{success rate} and \emph{coverage rate}. Success rate is the ratio of successful grasps among all predicted grasps. This metric only considers the grasp that is executed and does not contain any information about the other grasps. Predicting only one grasp is not suitable for 3D grasping, because the predicted grasp may lead to collision of the robot with other objects in the environment or there may not be any possible valid robot joint configuration that can reach the predicted grasp. In order to achieve an executable successful grasp, we need to generate a diverse set of grasps from different translations and directions to check for kinematic feasibility and collision avoidance.  As a result, we introduce the \emph{coverage rate} which captures the diversity of the grasps and measures how well the space of positive grasps $G^*$ is covered by the generated grasps. A positive grasp $g \in G^*$ is covered by the set of predicted grasps, $\hat{G}$, if there exists a grasp $\hat{g} \in \hat{G}$ that is at most \SI{2}{\cm} away from the grasp $g$. Positive grasps that have similar translation in object frame, have similar orientation. As a result, we chose to use the distance in translation of the grasps as the criterion for evaluating whether a grasp is covered or not. Since grasps are defined in $SE(3)$, $G^*$ is uncountably infinite. As a result, $G^*$ is approximated by sampling grasps while generating data.
 \emph{Success rate} and \emph{coverage rate} are analogous to precision and recall in the context of binary classification. Similar to precision-recall curves, we use the curves of \emph{success rate} and \emph{coverage rate} for analyzing and evaluating our method. We use the AUC of success-coverage rate for ablation studies and analysis.

\subsection{Analysis and Ablation Studies}
\label{sec:sim_experiments}

We evaluate the effect of different parameters and modules quantitatively using the same physics simulation as in the generation of training data~(Sec~\ref{sec:flex}).
For the ablation studies, we generate 86~object point cloud observations for 10~different objects that are held out during training. For each point cloud, 200~latent values are sampled and refined over 10 iterations, resulting in 2200 grasps per view point and 182,600 grasps in total.

\begin{figure}
    \centering
    \includegraphics[width=0.30\textwidth]{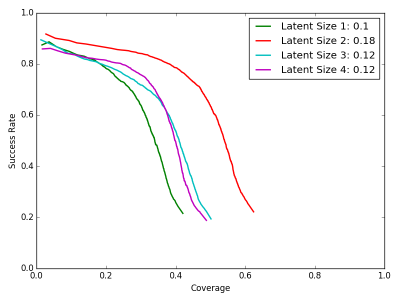}
    \caption{Effect of latent space dimensionality on success rate and coverage of the grasps.  Number in the box provide AUC values.}
    \label{fig:curve_pr}
\end{figure}

\paragraph{Dimensionality of the Latent Space} There is an inherent tension when deciding the dimensionality of the latent space which affects the quality of the generated grasps. The latent space needs to have enough capacity to allow the VAE to reconstruct the grasps. At the same time, a high-dimensional latent space leads to over-fitting and requires significantly more training data to be covered. It also deteriorates the quality of sampled grasps during inference, especially when the sampled latent values during inference are not seen by the generator network during training.
To analyze this effect, we evaluate the generator network with increasing numbers of dimensions for the latent space.
Fig.~\ref{fig:curve_pr} shows the resulting success-coverage curves for grasps generated at all the refinement iterations.
As can be seen, a dimensionality of one has the least AUC because the latent space does not have enough capacity. Although 3-dimensional and 4-dimensional latent spaces lead to a slightly better $\mathcal{L}_{vae}$ on the training data they perform worse during inference because the VAE cannot cover the latent space densely during training.
Given these results, we choose a two-dimensional latent space for all subsequent evaluations.

\begin{figure}
    \centering
    \includegraphics[width=0.30\textwidth]{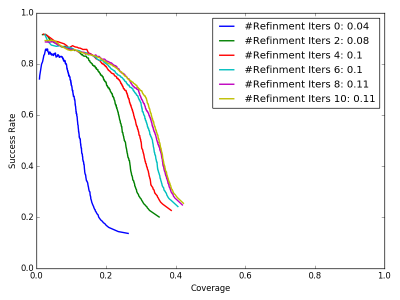}
    \caption{Effect of number of refinement steps on improving the accuracy and coverage of generated grasps. \ifarxiv Each curve is calculated over 16,600 grasps.\fi}
    \label{fig:curve_refinement}
\end{figure}

\paragraph{Effect of Refinement on Grasp Quality} While the grasp refinement increases the probability of success based on the evaluator network, it does not necessarily mean that the refined grasps succeed during test time. To analyze the actual improvement induced by each refinement step, we evaluate the grasps in simulation.
Fig.~\ref{fig:curve_refinement} shows the success-coverage curve of the grasps that are computed at each refinement iteration. As is shown, not only does the success rate of the generated grasps increase, the coverage rate increases as well. This is because when grasps are improved they get closer to the sampled positive grasps in $G^*$. The AUC of the curves plateaus after the 10th iteration of refinement. 

\begin{figure}
    \centering
    \ifarxiv
    \includegraphics[width=0.3\textwidth]{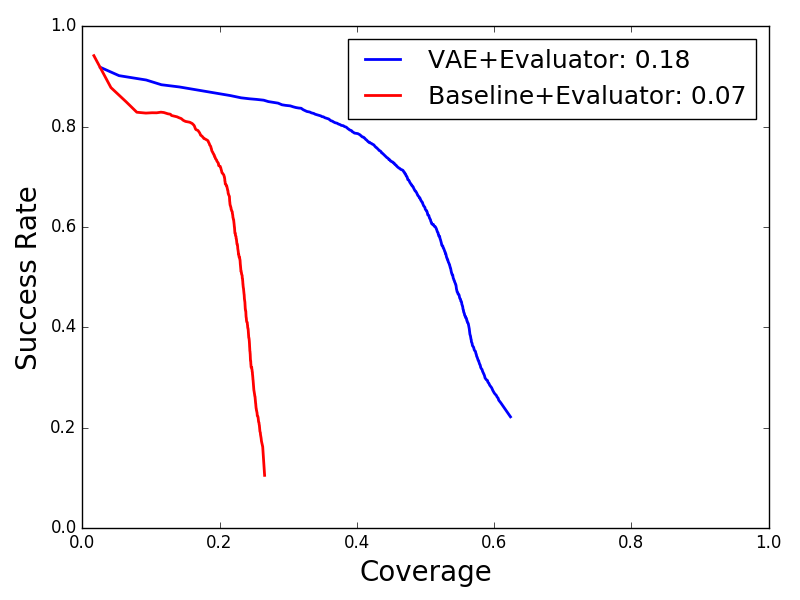}
    \caption{VAE Sampler vs. Geometric Sampler}
    \else
    \includegraphics[width=0.5\columnwidth]{figs/coverage_2000.png}%
    \includegraphics[width=0.5\columnwidth]{figs/rebuttal_pr_curve.png}
    \caption{\textit{Left:} Effect of number of sampled grasps on coverage rate. \textit{Right:} VAE Sampler vs. Geometric Sampler}
    \fi
    \label{fig:ablation}
\end{figure}

\paragraph{Effect of Sampled Grasps on Coverage} In previous sections, we conducted ablation studies using 200~random latent values because that was the maximum batch size that fits in GPU memory and it is the same setting that we used for our robot experiments. Consequently, the coverage rate in Fig.~\ref{fig:curve_refinement} was less than 0.5 even after 10 refinement steps. In order to investigate how the number of sampled grasps effect the coverage, 2000~grasps are sampled in 10~different batches on the same point clouds that were used in previous ablation studies. \ifarxiv Fig.~\ref{fig:2000} shows how more samples increase the coverage rate.
\begin{figure}
    \centering
    \includegraphics[width=0.30\textwidth]{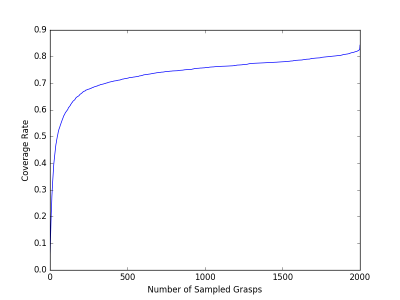}
    \caption{Effect of number of sampled grasps on coverage rate.}
    \label{fig:2000}
\end{figure}
\else
Fig.~\ref{fig:ablation} shows how more samples increase the coverage rate.
\fi

\paragraph{Learned Grasp Sampler vs. Geometric Grasp Sampler} In order to verify the effectiveness of using VAE for grasp sampling, we used the same geometric sampling method that was used to generate the training grasps. The baseline sampler estimates the surface normal from a point cloud and applies a random standoff and random planar rotation to the grasp and the evaluator takes the generated grasps, evaluates, and refines them. Fig.~\ref{fig:ablation} shows that VAE with latent size 2 is significantly better than the non-learning sampling scheme both for success and coverage. The surface normals rarely generate any grasps around the rims or thin structures. Also, this approach does not extrapolate to missing depths and occluded parts.

\subsection{Robot Experiments}
\label{sec:real_experiments}

\begin{table*}[]
    \centering
    \begin{tabular}{|c|c c c c|c|c|}
    \hline
         & Box & Cylinder & Bowl & Mug & Average Success Rate & Success Rate \\
         \hline
         6-DOF GraspNet & {\bf 83\%} & {\bf 89\%} & {\bf 100\%} & {\bf 86\%} & {\bf 90\%} & {\bf 88\%} \\
         GPD~\cite{ten2017grasp} &  50\% & 78\% & 78\% & 6\% & 52\% & 47\% \\
         \hline
    \end{tabular}
    \caption{Grasping results in real world experiments.}
    \label{tab:real_experiement}
\end{table*}

The ultimate test of the generated grasps is to execute them in the real world and deal with imperfect perception, robot joint limits, control errors, and physical phenomena such as friction that are difficult to model.
We want to show that: (1) our method scales to the real world despite being trained purely in simulation; (2) the generated grasp distribution is diverse enough to find successful grasps even after discarding those that violate robot kinematics and collision constraints; (3) our method's diverse grasp sampling leads to higher success rate in comparison to a state-of-the-art 6-DOF grasp planner~\cite{ten2017grasp}~(GPD).
\begin{figure}
    \centering
    \includegraphics[width=0.45\textwidth]{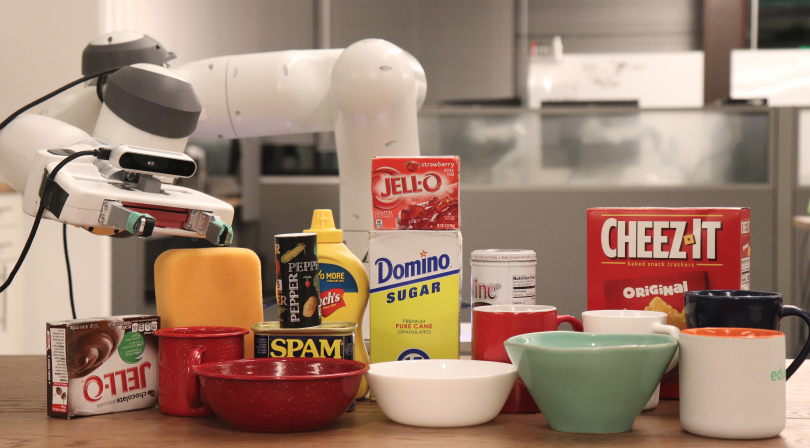}
    \caption{Each object is evaluated on three different poses. The 3D models of these objects are unknown. The training data consists of mugs, bowls, boxes, cylinder, and bottles with random scales. See the supplements for videos of the grasping trials.}
    \label{fig:robot_and_objects}
\end{figure}

All experiments are done using a 7-DOF~Franka Panda manipulator with an Intel RealSense D415 camera mounted on its parallel-jaw gripper. We choose a set of commonly-used objects that are challenging visually and physically. The weights of the objects are between $42g$ (pepper shaker) and $618g$ (mustard bottle). The hardware setup and object test set is shown in Fig.~\ref{fig:robot_and_objects}. The forward pass of VAE+Evaluator takes 0.04 seconds and each refinement iteration takes 0.3 seconds on batch size of 200 latent values using NVIDIA Titan XP.


 \paragraph{Protocol}
Each object is placed in three different stable poses on a table in front of the robot.
The robot's end-effector is moved such that the hand-mounted depth camera has an unobstructed view of the table top.
A grasp is considered successful if the robot can lift the object $10cm$ without dropping it. We filter the measured point cloud, remove the table plane and cluster the remaining points~\cite{PCL}.
This extracted object point cloud is the input to our approach and GPD.
Both methods return a list of scored grasps.
We use a motion planner to check for a collision-free path to each grasp pose and execute the one with the highest score.
If no grasp in the returned set can be executed we consider the trial a failure.
In total we run 51~trials per method.

\begin{figure}
    \centering
    \begin{tabular}{cc}
        \includegraphics[width=0.22\textwidth]{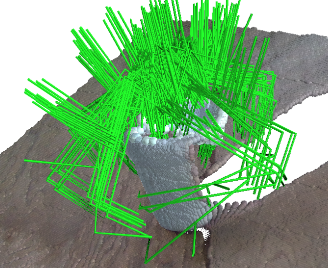} &  \includegraphics[width=0.22\textwidth]{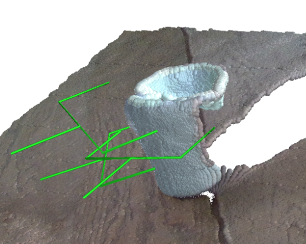} \\
    \end{tabular}
    \caption{Visualization of generated grasps by our method vs GPD~\cite{ten2017grasp} using 200 samples. (left) Generated grasps using 6-DOF GraspNet on a mug. (right) Generated grasps by GPD. Note that our method generates significantly more samples along the mug rim (and handle in other views). The object would slide out of the gripper for the side grasps.}
    \label{fig:gpd_sucks}
\end{figure}

\paragraph{Results}
Table~\ref{tab:real_experiement} shows that our method outperforms GPD~\cite{ten2017grasp} on success rate across all objects. One of the reasons is that our method generates diverse grasps which facilitates finding kinematically feasible ones. In contrast, GPD does not generate many different grasps which sometimes leads to situations in which no kinematically feasible grasp can be found. Mugs are particularly difficult for GPD because it does not generate any grasps from the rim (see Fig~\ref{fig:gpd_sucks}). \ifarxiv Another source of problem comes from the grasps where the finger gripper is tangent to one of the surfaces of the object. In these cases, slightest error in executing the grasp can change the grasp from grabbing the object to pushing it. 
The detailed outcomes of the experiments are shown in Table~\ref{tab:my_label}.
\fi

\ifarxiv
\begin{table*}[h]
    \centering
    \begin{tabular}{|c|c|c|c|c|c|c|c|c|c|}
    \hline
    
         \multirow{4}{*}{Categories} & \multirow{4}{*}{Objects} & \multicolumn{4}{c|}{6-DOF Grasp Net} & \multicolumn{4}{c|}{GPD~\cite{ten2017grasp}} \\\cline{3-10}
                                &   &  \multicolumn{3}{c|}{Trials} & \multirow{2}{*}{Success Rate} & \multicolumn{3}{c|}{Trials} & \multirow{2}{*}{Success Rate} \\\cline{3-5}\cline{7-9}
                                & &   \#1 & \#2 & \#3 & & \#1 & \#2 & \#3 & \\
         \hline
         \multirow{6}{*}{Box} & Jello Chocolate & \greencheck & \greencheck & \greencheck & 3/3 & \greencheck & \greencheck & \greencheck & 3/3 \\
         & Spam Meat Can & \greencheck & \redcross & \greencheck & 2/3 & \redcross & \greencheck & \magentacross & 1/3 \\
         & Jello Strawberry & \greencheck & \greencheck & \greencheck & 3/3 & \greencheck & \greencheck & \redcross & 2/3 \\
         & Sponge & \greencheck &\greencheck &\greencheck & 3/3 & \greencheck & \redcross & \redcross & 1/3 \\
         & Cheezit Box & \redcross & \redcross & \greencheck & 1/3 & \greencheck & \redcross & \redcross & 1/3 \\
         & Sugar Box & \greencheck &\greencheck &\greencheck &3/3 & \redcross & \redcross & \greencheck & 1/3 \\
         & Overall & & & & 15/18 & & & & 9/18 \\
         \hline
         \multirow{3}{*}{Cylinder} & Pepper Shaker & \greencheck & \greencheck & \redcross & 2/3 & \greencheck & \greencheck & \greencheck & 3/3 \\  
         & Piroluine & \greencheck &\greencheck &\greencheck &3/3 & \greencheck &\greencheck &\greencheck &3/3 \\
         & Mustard & \greencheck &\greencheck &\greencheck & 3/3 & \greencheck & \redcross & \redcross & 1/3 \\
         & Overall & & & & 8/9 & & & & 7/9 \\
         \hline
         \multirow{3}{*}{Bowl} & White Bowl & \greencheck & \greencheck & \greencheck & 3/3 & \magentacross & \redcross & \greencheck & 1/3 \\
         & Green Bowl & \greencheck &\greencheck &\greencheck &3/3 & \greencheck &\greencheck &\greencheck &3/3 \\
         & Red Bowl & \greencheck &\greencheck &\greencheck &3/3 & \greencheck &\greencheck &\greencheck &3/3 \\
         & Overall & & & & 9/9 & & & & 7/9 \\
         \hline
         \multirow{5}{*}{Mug} & White Mug & \greencheck &\greencheck &\greencheck & 3/3 & \magentacross & \magentacross & \redcross & 0/3 \\
         & Blue Mug & \greencheck & \greencheck & \redcross & 2/3 & \redcross & \magentacross & \redcross & 0/3 \\
         & Orange Mug & \greencheck &\greencheck &\greencheck & 3/3 & \redcross & \redcross & \greencheck & 1/3 \\
         & Big Red Mug & \redcross & \greencheck & \greencheck & 2/3 & \redcross & \redcross & \redcross & 0/3 \\
         & Red YCB Mug & \greencheck &\greencheck &\greencheck & 3/3 & \redcross & \magentacross & \redcross & 0/3 \\
         & Overall & & & & 13/15 & & & & 1/15 \\
         \hline
    \end{tabular}
    \caption{Detailed outcome of the robot experiments. \greencheck: the trial was successful, \redcross: generated grasp was not successful, \magentacross: none of the generated grasps are kinematically feasible.}
    \label{tab:my_label}
\end{table*}
\fi

%% file: 05_conclusions.tex
\section{Conclusions}
In this work, we introduced 6-DOF GraspNet for generating diverse grasps for unknown objects. Our method consists of a trained VAE that samples a variety of grasps for an object.  While the VAE is able to capture the complex distributions of successful grasp poses, it does not quite provide the accuracy required for highly robust grasp generation. To overcome this limitation, we additionally introduce a grasp evaluator network that assesses grasp quality and can refine grasps in an iterative process.  To the best of our knowledge, neither a learned grasp sampler nor a gradient-based refinement process have been introduced before. 

The training of our model is done using synthetic grasp data generated by a physics simulator. Therefore, our model can scale to large sets of objects without requiring the collection of data in the real world. We demonstrated that our method can transfer to the real world on objects with unknown 3D models by deploying the method on a real robot platform and an on-board RGB-D camera. We performed robot experiments on 17~objects with unknown 3D models and achieved state-of-the-art results in 3D grasping. We also performed a thorough analysis of the generated grasps in terms of success rate and coverage via ablation studies in a realistic physics simulator. 

This approach opens up a number of interesting directions in computer vision and robotics. In our method, all the latent values are sampled uniformly and then grasps are removed based on collision checks and kinematically feasible solutions. Potential extensions are to train the sampler or the evaluator in a way that not only considers the object of interest but also considers the surrounding objects to directly avoid generating colliding or infeasible grasps.  Other interesting directions are toward using the evaluator not only to refine sampled grasps but to provide real-time feedback guidance for a manipulator approaching an object. Our experiments provide evidence that our gradient-based approach could succeed in moving a manipulator closer and closer to a successful grasp.